\documentclass[12pt]{article}
\usepackage{graphicx}
\usepackage{epsfig}
\usepackage{amssymb}
\begin{document}
\title{SAGRAD: A Program for Neural Network Training with Simulated
Annealing and the Conjugate Gradient Method}
\author{Javier Bernal$^1$, Jose Torres-Jimenez$^{1,2}$ \\
$^1${\small \sl National Institute of Standards and Technology,} \\
{\small \sl Gaithersburg, MD 20899, USA} \\
$^2${\small \sl CINVESTAV-Tamaulipas, Information Technology Laboratory,} \\
{\small \sl Km. 5.5 Carretera Cd., Victoria, Tamaulipas, Mexico} \\
{\tt\small \{javier.bernal,jnt5\}@nist.gov \ \ jtj@cinvestav.mx}}
\date{\ }
\maketitle
\par
{\small \bf Abstract.}
{\small
SAGRAD (Simulated Annealing GRADient), a Fortran 77 program for computing neural
networks for classification
using batch learning, is discussed. Neural network training in SAGRAD is based
on a combination of simulated annealing and M\o ller's scaled conjugate gradient
algorithm, the latter a variation of the traditional conjugate gradient method,
better suited for the nonquadratic nature of neural networks. Different aspects
of the implementation of the training process in SAGRAD are discussed, such as
the efficient computation of gradients and multiplication of vectors by Hessian
matrices that are required by M\o ller's algorithm; the (re)initialization
of~weights with simulated annealing required to (re)start M\o ller's algorithm
the first time and each time thereafter that it shows insufficient progress in
reaching a possibly local minimum; and the use of simulated annealing when
M\o ller's algorithm, after possibly making considerable progress, becomes
stuck at a local minimum or flat area of weight space. Outlines of the scaled
conjugate gradient algorithm, the simulated annealing procedure and the
training process used in SAGRAD are presented together with results from
running SAGRAD on two examples of training data.
}\smallskip\par
\section{Introduction}
Neural networks are computational models that work by simulating the way
the brain processes information. They are often used to recognize patterns
in a data set, say~$X$, in Euclidean $d-$dimensional space, $d$~some
positive integer. Once the neural network is appropriately trained on
representative sample patterns of~$X$ , it can then be used for attempting
to recognize other patterns in~$X$ as they are fed through the~network.
Accordingly, it is assumed $X$ is partitioned into $n$ distinct
types/classes of patterns, $n$ some positive integer.
\par Let $A$ be a set of training data for~$X$, i.e., a subset of~$X$
in which the $n$ distinct types/classes of patterns are well represented.
The basic structure of a neural network associated with~$X$ (to be trained
on~$A$) consists of layers or columns of mostly computing nodes,
or neurons, arranged from left to right (see Figure~1) in such a way
that the result of a computation at each neuron in a layer contributes
to the input of neurons in the next layer.
\begin{figure}
\centerline{\includegraphics[angle=90]{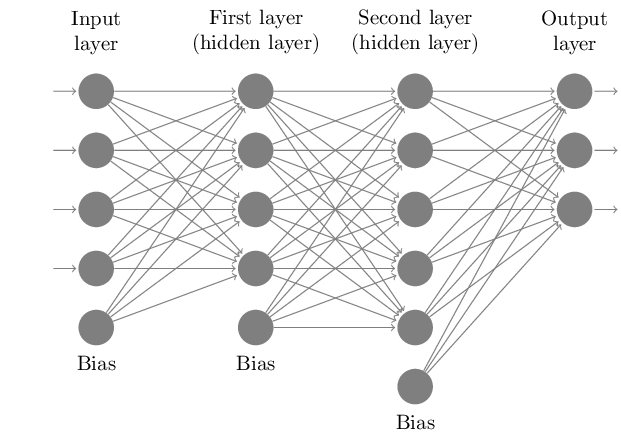}}
\caption{A $4-$layer neural network.}
\end{figure}
The layer at the extreme left of the network is called the input layer of the
network (see Figure~1) and consists of \mbox{$d+1$}~neurons.
A pattern vector in $X$, say \mbox{$a = \{a_k\}$}, \mbox{$k=1,\ldots,d$},
is introduced into the network through the input layer
as follows: $a$ is augmented to be of dimension~\mbox{$d+1$} by setting $a_{d+1}$
equal to~1; neurons in the input layer are labeled with integers from~1
to~\mbox{$d+1$}; and for each~$k$, \mbox{$k=1,\ldots,d+1$}, coordinate $a_k$ is
assigned to neuron~$k$ (neuron with label~$k$) and as such interpreted to be the
output of neuron~$k$ (neuron~$d+1$ is called a bias neuron
and its output is 1 for all patterns). The layer immediately to the right
of the input layer, unlike the input layer, consists of computing neurons
(except for the last neuron which is a bias neuron). From left to right in the
network it is the first layer with computing neurons and as such is
called the first layer of the network (the layer immediately to the right
of this layer is called the second layer of the network, and so on).
Like the input layer, the first layer has \mbox{$d+1$} neurons which are then labeled with
integers from \mbox{$d+2$} to~\mbox{$2d+2$}. Given integer~$i$,
\mbox{$d+2\leq i\leq 2d+1$}, a number $x_i$ is designated the input to neuron~$i$
(in the first layer) which is a weighted sum of the outputs of the input layer
(the coordinates of the augmented pattern~$a$) expressed as
\mbox{$x_i = \sum_{k=1}^{d+1} w_{ki}a_k$}.
Here for each $k$, \mbox{$k=1,\ldots,d+1$}, $w_{ki}$ is the weight modifying the
pattern coordinate $a_k$ before it is fed into neuron~$i$ (as part of~$x_i$).
In order to make neuron~$i$ into a computing neuron, the sigmoid activation
function \mbox{$\sigma(x)= 1/(1+e^{-x})$} is assigned to it. $\sigma$~is a
function with derivatives of all orders and values between 0 and~1.
\mbox{$y_i=\sigma(x_i)$}
is then designated the ouput of neuron~$i$, \mbox{$d+2\leq i\leq 2d+1$}, while
\mbox{$y_{2d+2} = 1$} is designated the output of neuron~\mbox{$2d+2$} (the
bias neuron). Inductively, given layers~$M$ and~$L$, consecutive layers in the
network from left to right; $\{y_m\}$, the set of outputs of neurons in layer~$M$;
$l_1$, $l_2$, \mbox{$l_1<l_2$}, integers such that neurons in layer~$L$ are
labeled with integers from $l_1$ to~$l_2$; and neuron~$l$, a neuron in layer~$L$,
\mbox{$l_1 \leq l \leq l_2-1$}; then a number $x_l$ is designated the input to
neuron~$l$ which is a weighted sum of the outputs of layer~$M$ expressed as
\mbox{$x_l = \sum_m w_{ml}y_m$}.
In addition the same sigmoid activation function $\sigma$ defined above is
assigned to neuron~$l$ and \mbox{$y_l=\sigma(x_l)$} is designated the output
of neuron~$l$, \mbox{$l_1\leq l\leq l_2-1$}, while \mbox{$y_{l_2} = 1$} is
designated the ouput of neuron~$l_2$ (the bias neuron of layer~$L$).
\par
The layer at the extreme right of the network is called the output layer of
the network (see Figure 1). Layers between the input layer and the output layer
are called hidden layers (in Figure~1 the first layer and second layer of the
network are the only hidden layers), and hidden layers to the right of the
first layer (there is only one, the second layer of the network, in the network
of Figure~1) are all assumed to be of the same length, i.e., to consist of the
same number of neurons, a number greater than~1 and preferably greater than
$d$ and~$n$. For consistency with definitions above involving consecutive
layers $M$ and $L$ we assume at first that the output layer contains a bias
neuron besides $n$ computing neurons.
As will become apparent below, there is a one-to-one correspondence between
the $n$ computing neurons in this layer and the $n$ classes of patterns
(as defined for $X$) into which the set $A$ of training data can be partitioned.
After reducing the number of neurons in the output layer to~$n$ by dropping the
dummy bias neuron in the layer, so that for some positive integer $nq$,
$nq$ is the total number of neurons in the network, neurons in the output
layer are then labeled with integers from \mbox{$nq-n+1$} to~$nq$.
Additionally, letting $nw$ be the total number of weights in the network,
a natural order can be established for weights so that any given set of $nw$
weights can be uniquely identified with a vector, called a weight
vector, in weight space, the Euclidean space of dimension~$nw$, and vice versa.
\par
Given a pattern $a$ in $A$, then for some $q$, \mbox{$1\leq q\leq n$}, $a$ is in
class~$q$, and an $n-$dimensional vector \mbox{$r(a)=\{r(a)_m\}$},
\mbox{$m=1,\ldots,n$}, called the desired response for~$a$, is defined by setting
$r(a)_q$ equal to~1 and $r(a)_m$ equal to~0 for \mbox{$m=1,\ldots,n$},~$m\not = q$.
Another $n-$dimensional vector \mbox{$o(a)=\{o(a)_m\}$}, \mbox{$m=1,\ldots,n$},
called the actual output for~$a$, is defined by setting $o(a)_m$ equal to the output
for $a$ of the $m^{th}$ neuron in the output layer (neuron with label $nq-n+m$)
for each $m$, \mbox{$m=1,\ldots,n$}. The (total) squared error between
desired responses $r(a)$ and actual outputs $o(a)$, $a$ in $A$, is then
\[ E(w)=1/2\sum_{a\in A}|r(a)-o(a)|^2 =
1/2\sum_{a\in A}\sum_{m=1}^n(r(a)_m -o(a)_m)^2,\]
where $w$ is the unique vector in weight space corresponding to the current set
of weights in the network. As $E$ is implicitly defined in terms of compositions
of linear functions between layers in the network and activation functions assigned
to neurons in the network, $E$ has partial derivatives of all orders at any~$w$.
Accordingly, any optimization method of the gradient kind can be applied
for the purpose of hopefully minimizing~$E$. If the result of training the neural
network on $A$, i.e., minimizing $E$ (with gradients, metaheuristics, etc.),
is a weight vector $w$ at which $E$ is zero then it must also be true that the neural
network defined by $w$ classifies correctly all patterns in the set $A$ of training
data, i.e., identifies correctly the class to which each pattern belongs. We say
then that $w$ is a reasonable solution. Additionally, if a subset of $X\setminus A$
is also available in which the $n$ distinct types/classes of patterns are also well
represented, and each pattern in the subset is of known classification, then the
neural network defined by $w$ should be applied on such a subset for classification
results.  If the results for a good percentage of the patterns in the subset,
say over 90~\%, are correct then we say that
besides being a reasonable solution, $w$ is also a quality solution.
\par In this paper we discuss SAGRAD, a Fortran 77 program for computing neural
networks for classification using batch learning. Classification is one of the
most important applications of neural networks. An extensive survey on neural
networks for classification can be found in \cite{zhang}. On the other hand,
batch learning is exactly the type of training described above where all patterns
in training data are introduced into the network before the training of the network
or minimization of the total error $E$ begins. This is in contrast with on-line
learning where training of the network is done one pattern at a time:
each time a pattern in the training data is introduced into the network, training
of the network takes place immediately starting at the current solution obtained
from introducing the previous pattern, and the training is done only on exactly those
patterns, including the current one, that have been introduced into the
network so far.
\par Neural network training in SAGRAD is based on a mixture of simulated
annealing~\cite{torr} and M\o ller's scaled conjugate gradient
algorithm~\cite{moll, moll3}, the latter a variation of the
traditional conjugate gradient method~\cite{hest}, better suited for the
nonquadratic nature of neural networks. In what follows an outline of M\o ller's
algorithm is presented that closely resembles the implementation of the algorithm
in SAGRAD. In addition, other aspects of the implementation of the training
process in SAGRAD are discussed such as the efficient computation of gradients
and multiplication of vectors by Hessian matrices that take place in M\o ller's
algorithm; the (re)initialization of~weights with simulated annealing required
to (re)start M\o ller's algorithm the first time and each time thereafter that
it shows insufficient progress in reaching a possibly local minimum; and the use
of simulated annealing when M\o ller's algorithm, after possibly making
considerable progress, becomes stuck at a local minimum or flat area of weight
space. Outlines of the simulated annealing procedure and the training process
used in SAGRAD are also presented together with results from training with
SAGRAD data for two examples.  A copy of \mbox{SAGRAD} can be found at
\verb+http://math.nist.gov/~JBernal+.
\section{Scaled Conjugate Gradient Algorithm}
SAGRAD is based on a combination of simulated annealing~\cite{torr} and
M\o ller's scaled conjugate gradient algorithm~\cite{moll, moll3}
for minimizing the total squared error $E$ as a function of weights.
M\o ller's algorithm, an outline of which is presented below,
is based on the well-known conjugate gradient method~\cite{hest} which works
well for quadratic or nearly-quadratic functions. Since the Hessian matrix $E''(w)$
of the squared error function $E$ at $w$ may not be positive definite for $w$ in
certain areas of weight space, M\o ller modified the conjugate
gradient method based on the approach of the Levenberg-Marquardt
algorithm~\cite{flet}. If at some point during the execution of the
conjugate gradient method for some $p$ and $w$ in $nw-$dimensional
Euclidean space \mbox{$\delta = p^t E''(w) p$} is computed resulting
in a nonpositive $\delta$, one makes $\delta$ positive by adding $p^t\lambda p$
to it for some~\mbox{$\lambda>0$}, i.e., by scaling the Hessian matrix $E''(w)$
with the appropriate~\mbox{$\lambda>0$} so that $\delta$ becomes
$p^t(E''(w)+\lambda I)p$, $I$ the identity matrix.
Once $\lambda$ is initialized it is used and
adjusted appropriately throughout the execution of the algorithm
so that each $\delta$ computed as above remains positive. However,
since the accuracy of the conjugate gradient method depends on
approximating $E(w)$ with a quadratic function that involves $E''(w)$, care
must be taken that the scaled $E''(w)$ does not produce a bad approximation.
This is again taken care of by appropriately raising and lowering~$\lambda$.
The outline of the scaled conjugate gradient algorithm below includes
the manipulations for raising and lowering~$\lambda$. Here the column vector
$E'(w)$ is the gradient of $E$ at weight vector~$w$. The outline closely
resembles the implementation of M\o ller's algorithm in~SAGRAD.
\begin{enumerate}
\item Initialize weight vector $w_0$,\\
$k=0$,
$\epsilon_1 = 10^{-6}$,
$\epsilon_2 = 10^{-4}$,
$\lambda_0 = \epsilon_2$,
$\bar{\lambda}_0=0$, $r_0 = p_0 = -E'(w_0)$,\\ $success=true$.
\item Calculate second-order information:
$s_k = E''(w_k)p_k$, $\delta_k = p_k^t s_k$.
\item Scale Hessian matrix:
$\delta_k = \delta_k + (\lambda_k -\bar{\lambda}_k)|p_k|^2$.
\item If $\delta_k \leq 0$ then scale Hessian matrix to make it positive
definite:\\
$\bar{\lambda}_k = 2(\lambda_k -\delta_k/|p_k|^2)$,
$\delta_k = -\delta_k +\lambda_k |p_k|^2$, $\lambda_k = \bar{\lambda}_k$.
\item Calculate the step size:
$\mu_k = p_k^t r_k$, $\alpha_k = \mu_k /\delta_k$.
\item Calculate the comparison parameter $\Delta_k$: $\bar{w}_{k+1} =
w_k+\alpha_k p_k$,\\ $E\_q = E(w_k) + E'(w_k)^t\alpha_k p_k +
1/2(\alpha_k p_k^t E''(w_k) \alpha_k p_k + \lambda_k |\alpha_k p_k|^2)$,\\
$\Delta_k = [E(w_k)-E(\bar{w}_{k+1})]/[E(w_k)-E\_q]
= 2 \delta_k [E(w_k)-E(\bar{w}_{k+1})]/\mu_k^2$.
\item Test for error reduction:\\
If $\Delta_k \geq 0$ then a successful error reduction
can be made:\\
$w_{k+1} = \bar{w}_{k+1}$, $r_{k+1} = -E'(w_{k+1})$.\\
If $|r_{k+1}|<\epsilon_1$ then terminate and return $w_{k+1}$ as
the desired minimum, perhaps not a global minimum.\\
If $success=false$ or $k \bmod nw = 0$ then restart:
$p_{k+1} = r_{k+1}$, $\lambda_{k+1} = \epsilon_2$, $\bar{\lambda}_{k+1} = 0$,
$k = k+1$, $success=true$, and go to step~2.\\
Else (if $success=true$ and $k \bmod nw \not = 0$) then create new
conjugate direction:
$\beta_k = (|r_{k+1}|^2 -r_{k+1}^t r_k)/\mu_k$,
$p_{k+1} = r_{k+1} + \beta_k p_k$.\\
If $\Delta_k \geq 0.75$ then reduce the scale parameter:
$\lambda_k = 1/2 \lambda_k$.\\
Else (if $\Delta_k < 0$) error reduction is not possible:\\
$\bar{\lambda}_k = \lambda_k$, $success=false$.
\item If $\Delta_k < 0.25$ then increase the scale parameter:
$\lambda_k = 4 \lambda_k$.
\item if $success=false$ then go to step 3.\\
Else set $\bar{\lambda}_{k+1}=0$, $\lambda_{k+1} = \lambda_k$,
$k=k+1$, and go to step 2.
\end{enumerate}
\section{Computing the Gradient}
In order to attempt to minimize the error $E$ as a function of $w$ using
the scaled conjugate gradient algorithm as described above, the capability
must exist for the efficient computation of the gradient $E'(w)$ of $E$ at~$w$
and multiplication of a vector by the Hessian matrix $E''(w)$ of $E$ at~$w$.
In this section we develop formulas used in SAGRAD for the
computation of the gradient $E'(w)$ as presented in~\cite{gonz}.
They originate from the so-called delta rule in \cite{rume}, \cite{widro}.
\par Given $a\in A$, $w$ in weight space, the error at $w$ due to $a$ is
$E_a(w) = 1/2\sum_{m=1}^n (r(a)_m-o(a)_m)^2$.
Thus, $E(w) = \sum_{a\in A}E_a(w)$.
Writing $w$ as $\{w_k\}$, \mbox{$k=1,\ldots,nw$}, it follows that
\mbox{$\partial E/\partial w_k = \sum_{a\in A} \partial E_a/\partial w_k$}
for each $k$, \mbox{$k=1,\ldots,nw$}. Therefore, by fixing $a$ in $A$, in what
follows it will suffice to develop only the formulas associated with $E_a$.
\par As will become apparent from the formulas below, the calculations of
the partial derivatives of $E_a$ with these formulas must take place in a
specific order, from right to left in the network. This is because each
calculation corresponding to a given weight depends on calculations
corresponding to other weights in the network to the right of the given
weight. Computing in this manner is called backpropagation, originally
described in~\cite{werbo}. However,
it is an implementation issue that is taken care of in SAGRAD and not
necessary for the development of the~formulas.
\par Consider layers $K$, $M$, $L$, consecutive layers in the network from
left to right, $\{y_k\}$, the set of outputs of neurons in layer~$K$,
$\{y_m\}$, the set of outputs of neurons in layer~$M$,
and $\{x_l\}$, the set of inputs of computing neurons in layer~$L$.
In particular consider $y_j$, the output of some neuron in layer~$K$,
and $x_i$, $y_i$, the input and output, respectively, of some
computing neuron in layer~$M$.
In addition, for each $y_k$ as above let $w_{ki}$ be the weight
such that \mbox{$x_i = \sum_k w_{ki}y_k$}; and for each $y_m$ and $x_l$ as
above let $w_{ml}$ be the weight such that \mbox{$x_l = \sum_m w_{ml}y_m$}.
\smallskip
\par Case 1. Layer $K$ is not the input layer and layer $M$ is a hidden
layer so that layer $L$ is either a hidden layer or the output layer.
\smallskip
\par Using the chain rule repeatedly we then get
\[\frac{\partial E_a}{\partial w_{ji}} =
\frac{\partial E_a}{\partial x_i}
\frac{\partial x_i}{\partial w_{ji}} =
\frac{\partial E_a}{\partial x_i}
\frac{\partial (\sum_k w_{ki}y_k)}{\partial w_{ji}} =
\frac{\partial E_a}{\partial x_i}
\frac{\partial (w_{ji}y_j)}{\partial w_{ji}} =
\frac{\partial E_a}{\partial x_i}y_j. \]
\[\frac{\partial E_a}{\partial x_i} =
\frac{\partial E_a}{\partial y_i}
\frac{\partial y_i}{\partial x_i} =
\frac{\partial E_a}{\partial y_i} \sigma'(x_i). \]
\[\frac{\partial E_a}{\partial y_i} =
\sum_l \frac{\partial E_a}{\partial x_l}
\frac{\partial x_l}{\partial y_i} =
\sum_l \frac{\partial E_a}{\partial x_l}
\frac{\partial (\sum_m w_{ml}y_m)}{\partial y_i} =
\sum_l \frac{\partial E_a}{\partial x_l}
\frac{\partial (w_{il}y_i)}{\partial y_i} = \]
\[\sum_l \frac{\partial E_a}{\partial x_l}w_{il}. \]
\par Thus,
\[\frac{\partial E_a}{\partial w_{ji}} =
\frac{\partial E_a}{\partial x_i}y_j =
\frac{\partial E_a}{\partial y_i} \sigma'(x_i)y_j =
(\sum_l \frac{\partial E_a}{\partial x_l}w_{il})\sigma'(x_i)y_j. \]
\par Case 2. Layer $K$ is the input layer so that layer $M$ is
the first layer in the network. Then $\{y_k\}$ can be replaced by~$\{a_k\}$,
the set of coordinates of input pattern~$a$.
\smallskip
\par Thus, $x_i = \sum_k w_{ki}a_k$, and
\[\frac{\partial E_a}{\partial w_{ji}} =
\frac{\partial E_a}{\partial x_i}a_j.\]
\par Once again
\[\frac{\partial E_a}{\partial x_i} =
\frac{\partial E_a}{\partial y_i} \sigma'(x_i) \]
and
\[\frac{\partial E_a}{\partial y_i} =
\sum_l \frac{\partial E_a}{\partial x_l}w_{il}. \]
\par Thus,
\[\frac{\partial E_a}{\partial w_{ji}} =
\frac{\partial E_a}{\partial x_i}a_j =
\frac{\partial E_a}{\partial y_i} \sigma'(x_i)a_j =
(\sum_l \frac{\partial E_a}{\partial x_l}w_{il})\sigma'(x_i)a_j. \]
\par Case 3. Layer $M$ is the output layer of the network so that there is
no layer~$L$.
\smallskip
\par Then once again
\[\frac{\partial E_a}{\partial w_{ji}} =
\frac{\partial E_a}{\partial x_i}y_j \]
and
\[\frac{\partial E_a}{\partial x_i} =
\frac{\partial E_a}{\partial y_i} \sigma'(x_i). \]
\par With $\{y_m\}$ ordered so that for $m=1,\ldots,n$, $y_m = o(a)_m$ then
$E_a(w) = 1/2\sum_{m=1}^n (r(a)_m-o(a)_m)^2 = 1/2\sum_{m=1}^n (r(a)_m-y_m)^2$,
so that
\[\frac{\partial E_a}{\partial y_i} =
y_i - r(a)_i\]
and
\[\frac{\partial E_a}{\partial w_{ji}} =
\frac{\partial E_a}{\partial x_i}y_j =
\frac{\partial E_a}{\partial y_i} \sigma'(x_i)y_j =
(y_i - r(a)_i)\sigma'(x_i)y_j. \]
\par Note that in all cases $\sigma'(x_i) = y_i (1-y_i)$.
\section{Fast Exact Multiplication by the Hessian}
In this section we develop the formulas used in the implementation of
the scaled conjugate gradient algorithm in SAGRAD for the fast
exact computation of the product of the Hessian matrix $E''(w)$ with
an $nw-$dimensional vector $v$ in the context of M\o ller's algorithm.
With these formulas the calculation of the complete Hessian matrix is
avoided. These formulas were originally derived by Pearlmutter \cite{pear}
and M\o ller \cite{moll2,moll3},
and involve the so-called $\mathcal{R}\{\cdot\}$ operator. As in the case
of the gradient $E'(w)$, by fixing $a$ in $A$, in what follows it will
suffice to develop only the formulas associated with $E_a$. These formulas
depend on the formulas developed above for the computation of the
gradient~$E'(w)$, thus simultaneously as~$E'(w)$ is computed with
backpropagation, the exact product of $v$ and~$E''(w)$ is computed
with these formulas in either a feed-forward fashion or in the manner of
backpropagation. But again this is an implementation issue that is taken
care of in SAGRAD and not necessary for the development of the~formulas.
\par Let $f$ be a differentiable function from $nw-$dimensional Euclidean
space into any other finite-dimensional Euclidean space.
The $\mathcal{R}_v\{\cdot\}$ operator or simply the $\mathcal{R}\{\cdot\}$
operator is defined by
\[ \mathcal{R}_v\{f(w)\}\equiv \frac{d}{dr}f(w+rv)\Bigg|_{r=0} = f'(w)v, \]
where $f'(w)$ is the Jacobian matrix of $f$ at~$w$. In particular
$\mathcal{R}_v\{E'(w)\} = E''(w)v$. Writing $w$ as $\{w_k\}$,
\mbox{$k=1,\ldots,nw$}, it also follows that for each $k$, \mbox{$k=1,\ldots,nw$},
\mbox{$\mathcal{R}_v\{\partial E/\partial w_k$\}} is the $k^{th}$ component
of~$E''(w)v$.
\par Given $g$, a differentiable function with domain and range appropriately
defined, $c$ a real number, then some equations involving
$\mathcal{R}\{\cdot\}$ are satisfied:
\begin{eqnarray*}
\mathcal{R}\{cf(w)\} & = & c\mathcal{R}\{f(w)\}\\
\mathcal{R}\{f(w)+g(w)\} & = & \mathcal{R}\{f(w)\}+\mathcal{R}\{g(w)\}\\
\mathcal{R}\{f(w)g(w)\} & = & \mathcal{R}\{f(w)\}g(w)+f(w)\mathcal{R}\{g(w)\}\\
\mathcal{R}\{g(f(w))\} & = & g'(f(w))\mathcal{R}\{f(w)\}\\
\mathcal{R}\{\frac{d}{dt}f(w)\} & = & \frac{d}{dt}\mathcal{R}\{f(w)\}\\
\mathcal{R}\{w\} & = & v.
\end{eqnarray*}
\par With these equations and the formulas obtained in the previous section
for the components of $E_a'(w)$, the formulas for the components of $E_a''(w)v$
can be derived. In what follows weights are doubly indexed. Since there is
a one-to-one correspondence between the components of $w$ and $v$ then
the components of $v$ will be similarly indexed.
\par As in the previous section, consider layers $K$, $M$, $L$,
consecutive layers in the network from left to right,
$\{y_k\}$, the set of outputs of neurons in layer~$K$,
$\{y_m\}$, the set of outputs of neurons in layer~$M$,
and $\{x_l\}$, the set of inputs of computing neurons in layer~$L$.
In particular consider $y_j$, the output of some neuron in layer~$K$,
and $x_i$, $y_i$, the input and output, respectively, of some computing
neuron in layer~$M$.
In addition, for each $y_k$ as above let $w_{ki}$ be the weight
such that \mbox{$x_i = \sum_k w_{ki}y_k$}; and for each $y_m$ and $x_l$ as
above let $w_{ml}$ be the weight such that \mbox{$x_l = \sum_m w_{ml}y_m$}.
\smallskip
\par Case 1. Layer $K$ is not the input layer and layer $M$ is a hidden
layer so that layer $L$ is either a hidden layer or the output layer.
\smallskip
\par Applying $\mathcal{R}\{\cdot\}$ on $x_i$ and $y_i$,
we get the feed-forward formulas:
\[ \mathcal{R}\{x_i\} = \mathcal{R}\{\sum_k w_{ki}y_k\} = \sum_k \mathcal{R}
\{w_{ki}y_k\} = \sum_k (\mathcal{R}\{w_{ki}\}y_k + w_{ki}\mathcal{R}\{y_k\}) = \]
\[ \sum_k (v_{ki}y_k + w_{ki}\mathcal{R}\{y_k\}). \]
\[ \mathcal{R}\{y_i\} = \mathcal{R}\{\sigma(x_i)\}=
\sigma'(x_i)\mathcal{R}\{x_i\}.\]
\par Applying $\mathcal{R}\{\cdot\}$ on $\partial E_a/\partial w_{ji}$,
$\partial E_a/\partial x_i$, $\partial E_a/\partial y_i$,
as computed in the previous section for case~1, we get the backpropagation
formulas:
\[ \mathcal{R}\{\frac{\partial E_a}{\partial w_{ji}}\} =
\mathcal{R}\{\frac{\partial E_a}{\partial x_i}y_j\} =
\mathcal{R}\{\frac{\partial E_a}{\partial x_i}\}y_j +
\frac{\partial E_a}{\partial x_i}\mathcal{R}\{y_j\}. \]
\[ \mathcal{R}\{\frac{\partial E_a}{\partial x_i}\} =
\mathcal{R}\{\frac{\partial E_a}{\partial y_i} \sigma'(x_i)\} =
\mathcal{R}\{\frac{\partial E_a}{\partial y_i}\} \sigma'(x_i) +
\frac{\partial E_a}{\partial y_i}\mathcal{R}\{\sigma'(x_i)\} = \]
\[ \mathcal{R}\{\frac{\partial E_a}{\partial y_i}\} \sigma'(x_i) +
\frac{\partial E_a}{\partial y_i}\sigma''(x_i)\mathcal{R}\{x_i\}. \]
\[ \mathcal{R}\{\frac{\partial E_a}{\partial y_i}\} =
\mathcal{R}\{\sum_l \frac{\partial E_a}{\partial x_l}w_{il}\} =
\sum_l \mathcal{R}\{\frac{\partial E_a}{\partial x_l}w_{il}\} =\]
\[\sum_l (\mathcal{R}\{\frac{\partial E_a}{\partial x_l}\}w_{il} +
\frac{\partial E_a}{\partial x_l}\mathcal{R}\{w_{il}\} =
\sum_l (\mathcal{R}\{\frac{\partial E_a}{\partial x_l}\}w_{il} +
\frac{\partial E_a}{\partial x_l}v_{il} ).
\]
\par Case 2. Layer $K$ is the input layer so that layer $M$ is
the first layer in the network. Then $\{y_k\}$ can be replaced by~$\{a_k\}$,
the set of coordinates of input pattern~$a$.
\smallskip
\par Applying $\mathcal{R}\{\cdot\}$ on $x_i$ and $\partial E_a/\partial w_{ji}$,
as computed in the previous section for case~2, we get
\[ \mathcal{R}\{x_i\} =
\mathcal{R}\{\sum_k w_{ki}a_k\} =
\sum_k \mathcal{R}\{w_{ki}a_k\} =
\sum_k \mathcal{R}\{w_{ki}\}a_k =
\sum_k v_{ki}a_k, \]
and
\[ \mathcal{R}\{\frac{\partial E_a}{\partial w_{ji}}\} =
\mathcal{R}\{\frac{\partial E_a}{\partial x_i}a_j\} =
\mathcal{R}\{\frac{\partial E_a}{\partial x_i}\}a_j, \]
with the other formulas derived for case 1 above remaining the same.
\smallskip
\par Case 3. Layer $M$ is the output layer of the network so that there is
no layer~$L$.
\smallskip
\par Applying $\mathcal{R}\{\cdot\}$ on $\partial E_a/\partial y_i$,
as computed in the previous section for case~3, we get
\[ \mathcal{R}\{\frac{\partial E_a}{\partial y_i}\} =
\mathcal{R}\{y_i-r(a)_i\} =
\mathcal{R}\{y_i\}, \]
with the other formulas derived for case 1 above remaining the same.
\smallskip
\par Note that in all cases $\sigma''(x_i) = (1-2y_i)\sigma'(x_i)$, and
$\sigma'(x_i) = y_i(1-y_i)$.
\section{Simulated Annealing}
An outline of the simulated annealing procedure used in SAGRAD follows.
It is based on the simulated annealing procedure presented in~\cite{torr}.
\par In the outline that follows $\epsilon$ is a tolerance reasonably chosen
(e.g., $\epsilon = 10^{-3}$). Accordingly if $w_b$ is the current best solution
found by the procedure and the squared error for $w_b$, i.e., $E_b=E(w_b)$, is
less than $\epsilon$ then $w_b$ is declared to be a reasonable solution and
the procedure is terminated.
\begin{enumerate}
\item Input: $w_i$, $nw$; $w_i$ a weight vector of $nw$ coordinates,
$nw>10$.
\item Initialize $K_1$, $K_2$, $temprture$, $tfactor$, $coef$, $\epsilon$ (e.g.,
$K_1=100$, $K_2=20$, $temprture~=~1.0$, $tfactor=0.99$, $coef=0.2$, $\epsilon =10^{-3}$).\\
Set $E_i=E(w_i)$, $E_b=E_i$, $w_c=w_i$, $E_c=\infty$, $k_2=0$, $k_0=0$, $k_c=0$.\\
Initialize $nb$ to a positive integer relatively small with respect
to~$nw$ (e.g., $nb=$ largest integer $\leq 0.05\cdot nw$ if $0.05\cdot nw \geq 2$,
$nb=2$ otherwise).
\item If $k_c=k_0$ then set $temprture = tfactor\cdot temprture$.\\
Set $k_1=0$, $k_2 = k_2+1$, $k_c=k_0$.
\item Set $k_1 = k_1+1$, $w_f=w_c$.\\
For $n=1,\ldots,nw$, let $w_f(n)$ be the $n^{th}$ coordinate of $w_f$.\\
Generate random integer $nv$, $1\leq nv\leq nb$.\\
Generate distinct random integers $j_m$, $m=1,\ldots,nv$,
$1\leq j_m\leq nw$.\\
For each $m$, $m=1,\ldots,nv$, generate random number $r(m)$ in
$(-1,1)$, and set $w_f(j_m)=w_f(j_m) + coef\cdot r(m)$.\\
Set $E_f=E(w_f)$.\\
If $E_f<E_b$ then set $w_b=w_f$, $w_c=w_f$, $E_b=E_f$, $E_c=E_f$,
$k_0=k_0+1$.\\
Else if $E_f<E_c$ then set $w_c=w_f$, $E_c=E_f$.\\
Else if $E_f\geq E_c$ then generate random number $r$ in $(0,1)$,\\
and if $r<\exp ^{(E_c-E_f)/temprture}$ then set $w_c=w_f$, $E_c=E_f$.
\item If $k_1 < K_1$ then go to step 4.
\item If $k_0>0$ and $E_b<\epsilon$ then go to step 8.
\item If $k_2 < K_2$ then go to step 3.
\item If $k_0>0$ then set $w_c=w_b$, $E_c=E_b$ (the input solution
$w_i$ was improved and $w_c$ now equals the best solution found by
the procedure).
\item Output: $w_c$, $E_c$, $k_0$ (if $k_0>0$ then the input solution
$w_i$ was improved and $w_c$ equals the best solution found by the
procedure; if $k_0=0$ then the initial solution $w_i$ was not improved
and $w_c$ equals the last solution that was not an improvement but
was still accepted by the procedure).
\end{enumerate}
\par Two versions of the simulated annealing procedure outlined above
are used in the training process in SAGRAD. One of low intensity with
a relatively small number of iterations, high initial temperature
and small neighborhood of exploration (currently with $K_1=100$,
$K_2=20$, $temprture=1.0$, $tfactor=0.99$, $coef=0.2$,
$\epsilon=10^{-3}$), and one of high intensity with a relatively
large number of iterations, low initial temperature and large
neighborhood of exploration (currently with $K_1=5000$, $K_2=250$,
$temprture=0.1$, $tfactor=0.99$, $coef=1.0$, $\epsilon=10^{-3}$).
As pointed out in~\cite{lede}, neural network training is typically
a two-step process. First, with a method such as the low-intensity
version of the simulated annealing procedure mentioned above that tends
to elude local minima, weights are initialized. Then an optimization
algorithm such as the conjugate gradient method is applied in hopes of
finding a global~minimum.
\par In the training process in SAGRAD we follow a slight variation of
this two-step strategy while adding an additional step. The additional
step involves the high-intensity version of the simulated annealing
procedure mentioned above that intensively exploits weight space for
a possible global solution. It is used principally when the scaled
conjugate gradient algorithm (in the second step), after possibly making
considerable progress, becomes stuck at a local minimum or flat area
of weight space.
\section{Training Process}
In SAGRAD, neural network training is essentially a
three-step process that while still following the two-step strategy in
\cite{lede}, combines in a slightly different manner the two versions
of the simulated annealing procedure mentioned above with M\o ller's
scaled conjugate gradient algorithm. An outline of the training process
in SAGRAD in terms of the three steps follows below. There and in what
follows a weight vector will be declared to be a reasonable solution
if for some reasonably chosen $\epsilon$ (e.g.,~$\epsilon = 10^{-3}$),
the squared error for the weight vector is less than~$\epsilon$.
It should also be noted that at different
times during the execution of the process, in order to provide the
user with the option of getting out of a possibly bad run of the
training process, the user will be asked to decide on whether or not to
terminate the current run of the training process. If the run is terminated
then the user will be asked to decide on whether SAGRAD should stop or do a
new cold start of the training process.
\par In the outline of the training process below note that before the
third step is executed (at most once), the first two steps, one following
the other, may be executed several times in hopes that the scaled conjugate
gradient algorithm in the second step will eventually compute a reasonable
solution. The scaled conjugate gradient algorithm in the second step uses
as its initial solution the output weights from the last execution of the
low-intensity simulated annealing in the first step. On the other hand,
the low-intensity simulated annealing in the first step uses as input the
best weights found so far among all executions of the scaled conjugate
gradient algorithm in the second step except at the start of the execution
of the process. At the start of the execution of the process the
low-intensity simulated annealing in the first step uses input weights
that are randomly generated in the interval~$(-1,1)$. It should be
noted that all executions of the low-intesity simulated annealing in the
first step tend to produce good initial solutions for the scaled conjugate
gradient algorithm in the second step. However it is the first
execution of the low-intensity simulated annealing in the first step that
usually reduces considerably the squared error while all others usually
produce no reduction at all. Eventually, as the first two steps are
repeatedly executed, either a reasonable solution is found and the process
is terminated, or the third step, which involves an execution of the
high-intensity simulated annealing followed by an execution of the scaled
conjugate gradient algorithm, is executed one time in hopes of finding
a reasonable solution. Note that even if a reasonable solution has been found
by executing only the first two steps, the user can still direct the training
process to go to the third step for a perhaps better solution. Note as well
that it may take several cold starts of the training process before a
reasonable solution is obtained.
\begin{enumerate}
\item (Using low-intensity simulated annealing procedure)\\
If at start of execution of process then generate weight vector
$w_i$ with coordinates random numbers in the interval~$(-1,1)$;
set~$k_3~=~1$.\\
Otherwise step 2 below and this step have been executed previously:\\
set $w_i=w_m$, where $w_m$ is the best solution found so far among all executions
of the scaled conjugate gradient algorithm in step~2 below; set $k_3 = k_3+1$.\\
With $w_i$ as the input weight vector, execute simulated annealing
procedure in previous section using its low-intensity version to produce
output weight vector~$w_c$.
\item (Using scaled conjugate gradient algorithm)\\
Execute scaled conjugate gradient algorithm with weight vector $w_0$ in
step 1 of the outline of the algorithm in Section~2 initialized at $w_c$.\\
At any time during the execution of the algorithm, let $w_m$ be the best solution
found so far among all executions in this step, including
current one, of the scaled conjugate gradient algorithm.\\
At any time if $w_m$ is a reasonable solution then terminate the training process
(unless the user directs the training process to go to step~3 below
for a perhaps better solution).\\
At any time if ``$|r_{k+1}|<\epsilon_1$'' in step~7 of the outline of the algorithm
in Section~2 has not occurred (note $r_{k+1}$ is the negative of the gradient of
the squared error function $E$ at the current solution~$w_{k+1}$) and $k>iter$
(e.g, $iter = 10\cdot nw$, $nw$ the number of weights in the network) then if
$k_3<K_3$ (e.g., $K_3=20$), then go to step~1 above. Otherwise ($k_3\geq K_3$)
go to step~3 below.\\
At any time if ``$|r_{k+1}|<\epsilon_1$'' in step~7 of the outline of the algorithm
in Section~2 has occurred and since $w_m$ is not a reasonable solution so that
algorithm is possibly stuck at either a local minimum, i.e., current solution
$w_{k+1}$, or flat area of weight space, then go to step~3 below.
\item (Using high-intensity simulated annealing procedure and
scaled conjugate gradient algorithm)\\
Set $w_i=w_m$, where $w_m$ is the best solution found so far among all executions
of the scaled conjugate gradient algorithm in step~2 above.\\
With $w_i$ as the input weight vector, execute simulated annealing
procedure in previous section using its high-intensity version to produce
output weight vector~$w_c$.\\
Execute scaled conjugate gradient algorithm with weight vector $w_0$ in
step 1 of the outline of the algorithm in Section~2 initialized at $w_c$.\\
At any time during the execution of the algorithm, let $w_m$ be the best solution
found so far among this execution and all executions in step 2 above of the
scaled conjugate gradient algorithm.\\
At any time if $w_m$ is a reasonable solution, or if ``$|r_{k+1}|<\epsilon_1$''
in step~7 of the outline of the algorithm in Section~2 has not occurred and
$k>iter$, or if ``$|r_{k+1}|<\epsilon_1$'' in step~7 of the outline of the algorithm
in Section~2 has occurred, then terminate executions of the scaled conjugate gradient
algorithm and the training process. Return $w_m$ as the best~solution.
\end{enumerate}
\par If $w_m$ is not a reasonable solution then the user should
direct SAGRAD to do at least one more cold start of the above training
process. Even if $w_m$ is a reasonable solution the user can always direct
SAGRAD to do more cold starts of the training proces in hopes of getting a
better solution. 
\section{Numerical Results}
\subsection{Cushing Syndrome Classification}
Here we present results from running SAGRAD on a small
example associated with the so-called Cushing syndrome. This is an
example used in \cite{gong} as an application of neural networks for
classification. \par The Cushing syndrome is a disorder that occurs when
the body is exposed to high levels of the hormone cortisol for
a long time. Three types of the syndrome are recognized:
adenoma, bilateral hyperplasia, and carcinoma. In the presence
of the Cushing syndrome the following observations were
made that represent urinary excretion rates (mg/24h) of the
steroid metabolites tetrahydrocortisone (in the second column below)
and pregnanetriol (in the third column). Each line of observations has
a label that appears in the first column, and each of the lines
corresponds to an individual identified with each of the observations in
the line, an individual with a known type of the syndrome. Accordingly,
lines labeled a1, ..., a6 correspond to individuals with
the adenoma type; lines labeled b1, ..., b10 correspond to individuals with
the bilateral hyperplasia type; and lines labeled c1, ..., c5 correspond
to individuals with the carcinoma type. Lines labeled u1, ..., u6 correspond
to individuals with the syndrome, each individual with an unknown type of
the syndrome. Finally, the fourth and fifth columns of the data below have
the same data as the second and third columns, respectively,
but on a $\log$ scale.
\\ \smallskip\\
a1 \ \ \ \        3.1 \          11.70 \ \ \    1.1314021 \ \ \ \ 2.45958884\\
a2 \ \ \ \        3.0 \ \ \       1.30 \ \ \    1.0986123 \ \ \ \ 0.26236426\\
a3 \ \ \ \        1.9 \ \ \       0.10 \ \ \    0.6418539 \ \ \  -2.30258509\\
a4 \ \ \ \        3.8 \ \ \       0.04 \ \ \    1.3350011 \ \ \  -3.21887582\\
a5 \ \ \ \        4.1 \ \ \       1.10 \ \ \    1.4109870 \ \ \ \ 0.09531018\\
a6 \ \ \ \        1.9 \ \ \       0.40 \ \ \    0.6418539 \ \ \  -0.91629073\\
b1 \ \ \ \        8.3 \ \ \       1.00 \ \ \    2.1162555 \ \ \ \ 0.00000000\\
b2 \ \ \ \        3.8 \ \ \       0.20 \ \ \    1.3350011 \ \ \  -1.60943791\\
b3 \ \ \ \        3.9 \ \ \       0.60 \ \ \    1.3609766 \ \ \  -0.51082562\\
b4 \ \ \ \        7.8 \ \ \       1.20 \ \ \    2.0541237 \ \ \ \ 0.18232156\\
b5 \ \ \ \        9.1 \ \ \       0.60 \ \ \    2.2082744 \ \ \  -0.51082562\\
b6 \ \ \         15.4 \ \ \       3.60 \ \ \    2.7343675 \ \ \ \ 1.28093385\\
b7 \ \ \ \        7.7 \ \ \       1.60 \ \ \    2.0412203 \ \ \ \ 0.47000363\\
b8 \ \ \ \        6.5 \ \ \       0.40 \ \ \    1.8718022 \ \ \  -0.91629073\\
b9 \ \ \ \        5.7 \ \ \       0.40 \ \ \    1.7404662 \ \ \  -0.91629073\\
b10  \           13.6 \ \ \       1.60 \ \ \    2.6100698 \ \ \ \ 0.47000363\\
c1 \ \ \         10.2 \ \ \       6.40 \ \ \    2.3223877 \ \ \ \ 1.85629799\\
c2 \ \ \ \        9.2 \ \ \       7.90 \ \ \    2.2192035 \ \ \ \ 2.06686276\\
c3 \ \ \ \        9.6 \ \ \       3.10 \ \ \    2.2617631 \ \ \ \ 1.13140211\\
c4 \ \ \         53.8 \ \ \       2.50 \ \ \    3.9852735 \ \ \ \ 0.91629073\\
c5 \ \ \         15.8 \ \ \       7.60 \ \ \    2.7600099 \ \ \ \ 2.02814825\\
u1 \ \ \ \        5.1 \ \ \       0.40 \ \ \    1.6292405 \ \ \  -0.9162907\\
u2 \ \ \         12.9 \ \ \       5.00 \ \ \    2.5572273 \ \ \ \ 1.6094379\\
u3 \ \ \         13.0 \ \ \       0.80 \ \ \    2.5649494 \ \ \  -0.2231436\\
u4 \ \ \ \        2.6 \ \ \       0.10 \ \ \    0.9555114 \ \ \  -2.3025851\\
u5 \ \ \         30.0 \ \ \       0.10 \ \ \    3.4011974 \ \ \  -2.3025851\\
u6 \ \ \         20.5 \ \ \       0.80 \ \ \    3.0204249 \ \ \  -0.2231436\\
\smallskip \\
Log scale data above for observations in lines a1, ..., a6, b1, ..., b10,
c1, ..., c5, was used as training data for SAGRAD, and after only one cold
start of training process, training was
completed on a $4-$layer network associated with the data. The input layer
of this network had 3 nodes, the first layer 3, the second layer 4, and
the ouput layer 3. Then log scale data for observations in lines u1, ..., u6,
together with the trained network was used to identify with SAGRAD the type
(adenoma, bilateral hyperplasia, or carcinoma) corresponding to each
of these lines. The classification results from the execution of
SAGRAD follow for each line of unknown type.
Here the first columm of numbers contains outputs from the ouput node
of the neural network corresponding to the ademona type; the second
column contains outputs from the output node corresponding to the
bilateral hyperplasia type; and finally the third column contains
outputs from the output node corresponding to the carcinoma type. 
\\ \smallskip \\
u1 \ \ \  1.57569381E-25 \ \ \  1.\ \ \ \ \ \ \ \ \ \ \ \ \ \ \ \ \ \ \ \ \ \  3.27244446E-61\\
u2 \ \ \  1.71192768E-10 \ \ \  4.88008365E-06 \ \ \  0.999999999\\
u3 \ \ \  1.07917885E-38 \ \ \  1.\ \ \ \ \ \ \ \ \ \ \ \ \ \ \ \ \ \ \ \ \ \  2.25400938E-41\\
u4 \ \ \  0.99999641 \ \ \ \ \ \ \ \ \  1.47077866E-05 \ \ \  7.78196988E-08\\
u5 \ \ \  1.04024021E-38 \ \ \  1.\ \ \ \ \ \ \ \ \ \ \ \ \ \ \ \ \ \ \ \ \ \  2.11941796E-41\\
u6 \ \ \  1.21287938E-38 \ \ \  1.\ \ \ \ \ \ \ \ \ \ \ \ \ \ \ \ \ \ \ \ \ \  2.77865859E-41
\\ \smallskip \\
From these results it appears that adenoma is the type corresponding to line u4;
bilateral hyperplasia is the type corresponding to lines u1, u3, u5, u6;
and carcinoma is the type corresponding to line~u2. This classification of
these lines is consistent with the classification of the same lines
in~\cite{gong}.

\subsection{Wine Classification}
Data in \cite{bach} is the result of a chemical analysis of wines
produced in the same region in Italy from three different cultivars.
Each line in the data corresponds to a wine and contains quantities
of 13 constituents in the wine that were determined through the
chemical analysis.
\par The 13 constituents were:\\
1) Alcohol\\
2) Malic acid\\
3) Ash\\
4) Alkalinity of ash\\
5) Magnesium\\
6) Total phenols\\
7) Flavanoids\\
8) Nonflavanoid phenols\\
9) Proanthocyanins\\
10)Color intensity\\
11)Hue\\
12)OD280/OD315 of diluted wines\\
13)Proline \smallskip
\par
Training data for SAGRAD was obtained from \cite{bach} as
follows. The first 50 lines of data for wine from the first cultivar
were extracted from the data and identified as Class 1 training data;
the first 60 lines of data for wine from the second cultivar
were extracted from the data and identified as Class 2 training data;
and the first 40 lines of data for wine from the third cultivar
were extracted from the data and identified as Class 3 training data.
In all cases each line of data consisted of 13 numbers
corresponding in the same order to the quantities of constituents
listed above. For example, the first line in the Class 1 training
data appeared exactly as follows:\\ \smallskip\\
4.23,1.71,2.43,15.6,127,2.8,3.06,.28,2.29,5.64,1.04,3.92,1065\\
\smallskip
\par
Using this data, SAGRAD was then executed, and after two cold starts of
training process, training was completed on a $4-$layer network
associated with the data. The input layer of this network had 14 nodes,
the first layer had 14, the second layer had 15, and the output layer
had 3. For the purpose of testing the trained
network the remaining 9 lines of data for wine
from the first cultivar were extracted from the data and
identified as Class 1 independent data; the remaining 11 lines
of data for wine from the second cultivar were extracted from
the data and identified as Class 2 independent data; and the
remaining 8 lines of data for wine from the third cultivar were
extracted from the data and identified as Class 3 independent data.
\par
The classification results from the execution of SAGRAD follow for each
line of independent data. Here the first columm of numbers contains
outputs from the ouput node of the neural network corresponding to
wine from the first cultivar; the second column contains outputs from
the output node corresponding to wine from the second cultivar; and
finally the third column contains outputs from the output node
corresponding to wine from the third cultivar. The first 9 lines
correspond to the 9 lines in the Class 1 independent data in the
same order; the next 11 lines correspond to the 11 lines in the
Class 2 independent data in the same order; and the final 8 lines  
correspond to the 8 lines in the Class 3 independent data in the
same order.\\ \smallskip\\
  0.999999748  1.4638003E-10  3.87972262E-19\\
  0.999999748  1.4638003E-10  3.87972262E-19\\
  0.999999748  1.4638003E-10  3.87972262E-19\\
  0.999999748  1.4638003E-10  3.87972262E-19\\
  0.999999748  1.4638003E-10  3.87972262E-19\\
  0.999999748  1.4638003E-10  3.87972262E-19\\
  0.999999748  1.4638003E-10  3.87972262E-19\\
  0.999999748  1.4638003E-10  3.87972262E-19\\
  0.999999748  1.4638003E-10  3.87972262E-19\\
  0.00725871591  1.  2.44696739E-57\\
  0.999999746  1.47618842E-10  3.86290537E-19\\
  0.00725899153  1.  2.44597986E-57\\
  0.00725948808  1.  2.44420191E-57\\
  0.00725610139  1.  2.45635659E-57\\
  0.0072496458  1.  2.47970923E-57\\
  0.00725948807  1.  2.44420193E-57\\
  0.00725948808  1.  2.44420191E-57\\
  0.00725948808  1.  2.44420191E-57\\
  0.00725948808  1.  2.44420191E-57\\
  3.35969945E-05  1.  1.41314074E-32\\
  2.0606421E-10  5.70930171E-17  1.\\
  2.06064211E-10  5.7093017E-17  1.\\
  2.06064208E-10  5.70930172E-17  1.\\
  2.06064208E-10  5.70930172E-17  1.\\
  2.06065502E-10  5.70929678E-17  1.\\
  2.06064208E-10  5.70930172E-17  1.\\
  2.06064208E-10  5.70930172E-17  1.\\
  2.54410205E-19  1.07524957E-06  1.\\ \smallskip \\
From these results it appears that only the wine corresponding
to the $2^{nd}$ line in the Class 2 independent data was
classified incorrectly. Additional output from SAGRAD confirms this:
\\ \smallskip\\
 Independent patterns classification results:\\
 Class =  1 Total =  9 Correct =  9 Percentage =   100.\\
 Class =  2 Total =  11 Correct =  10 Percentage =   90.9090909\\
 Class =  3 Total =  8 Correct =  8 Percentage =   100.0\\ \smallskip \\
These results compare well with results found elsewhere for the same
wine data, e.g., in \cite{niko}, \cite{ozgur}, \cite{stojk}.
\section{Summary}
SAGRAD, a Fortran 77 program for computing neural networks for classification
using batch learning, was discussed. Since neural network training in SAGRAD
is based in part on M\o ller's scaled conjugate gradient algorithm which is a
variation of the traditional conjugate gradient method, better suited for the
nonquadratic nature of neural networks, an outline of M\o ller's algorithm was
presented that resembles its implementation in SAGRAD. Important aspects of
the implementation of the training process in SAGRAD were discussed such as
the efficient computation of gradients and multiplication of vectors by
Hessian matrices that are required by M\o ller's algorithm. Accordingly,
formulas for the product of vectors by Hessian matrices depending on those
for the gradients used in SAGRAD were developed. Because of this dependence
it was pointed out that calculations with these formulas of the gradient at
a vector and the product of the Hessian at the same vector with another
vector in the context of M\o ller's algorithm occur simultaneously and take
place in either a feed-forward fashion or in the manner of backpropagation.
\par As neural network training in SAGRAD is also based on simulated annealing,
an outline of the simulated annealing procedure implemented in SAGRAD was also
presented. It was then pointed out that two versions of this procedure are used
in the training process in SAGRAD, one a low-intensity version for the
(re)initialization of weights required to (re)start the scaled conjugate
gradient algorithm the first time and each time thereafter that it shows
insufficient progress in reaching a possibly local minimum; and the other
a high-intensity version to be used once the scaled conjugate gradient
algorithm has possibly reduced the squared error considerably but becomes
stuck at a local minimum or flat area of weight~space. An outline of the
training process was then presented.
\par Finally, results from executions of SAGRAD were reported. SAGRAD was
run on two essentially small examples of training data consisting of sample
patterns of dimension 2 and~13, respectively. The trainings with SAGRAD of the
training data for the two examples were declared to be good as reasonable
solutions were obtained. Classification results were then presented from
applying the corresponding neural networks on the trained data, and other
independent data of known and unknown classification, and these results
were also declared to be good as for over 90~\% of the patterns in the
data the results were correct.
\par It should be noted that in general it may take several cold starts
of the training process in SAGRAD before a reasonable solution is obtained.
Such a solution should then be tested for quality by applying the corresponding
neural network on independent sample patterns of known classification.
A~copy of \mbox{SAGRAD} can be obtained at
\verb+http://math.nist.gov/~JBernal+.

\end{document}